%% file: root.tex
\documentclass[letterpaper, 10 pt, conference]{ieeeconf}  

\IEEEoverridecommandlockouts                              

\input{macros.tex}
\title{\LARGE \bf Using Lidar Intensity for Robot Navigation }

\author{Adarsh Jagan Sathyamoorthy$^{1*}$, Kasun Weerakoon$^{1*}$, Mohamed Elnoor$^{1}$, and Dinesh Manocha$^{1}$.
\thanks{$^{*}$ Authors contributed equally.}}

\begin{document}

\maketitle
\thispagestyle{empty}
\pagestyle{empty}

\footnotetext[1]{Authors are with the University of Maryland, College Park.}


\begin{abstract}
We present Multi-Layer Intensity Map, a novel 3D object representation for robot perception and autonomous navigation. Intensity maps consist of multiple stacked layers of 2D grid maps each derived from reflected point cloud intensities corresponding to a certain height interval. The different layers of intensity maps can be used to simultaneously estimate obstacles' height, solidity/density, and opacity. We demonstrate that intensity maps' can help accurately differentiate obstacles that are safe to navigate through (e.g. beaded/string curtains, pliable tall grass), from ones that must be avoided (e.g. transparent surfaces such as glass walls, bushes, trees, etc.) in indoor and outdoor environments. Further, to handle narrow passages, and navigate through non-solid obstacles in dense environments, we propose an approach to adaptively inflate or enlarge the obstacles detected on intensity maps based on their solidity, and the robot's preferred velocity direction. We demonstrate these improved navigation capabilities in real-world narrow, dense environments using a real Turtlebot and Boston Dynamics Spot robots. We observe significant increases in success rates to more than 50\%, up to a 9.5\% decrease in normalized trajectory length, and up to a 22.6\% increase in the F-score compared to current navigation methods using other sensor modalities. 
\end{abstract}


\input{1_Introduction}
\input{2_Related_Work}

\input{3_Background}
\input{4_Intensity_Map}
\input{5_Results}

\input{6_Conclusions}

\bibliographystyle{IEEEtran}
\bibliography{References}

\end{document}

%% file: macros.tex
\usepackage{xcolor}
\usepackage{amssymb}
\usepackage{graphicx}
\usepackage{mathtools}
\usepackage{amsmath}
\usepackage{gensymb}
\usepackage{float}
\usepackage{multirow}
\usepackage{makecell}
\usepackage{mathtools}
\usepackage{hyperref}
\hypersetup{
    colorlinks=true,
    linkcolor=blue,
    filecolor=magenta,      
    urlcolor=blue,
    pdftitle={Overleaf Example},
    pdfpagemode=FullScreen,
    }

\usepackage{tabularx}
    \newcolumntype{L}{>{\raggedright\arraybackslash}X}
\usepackage[utf8]{inputenc}
\usepackage[english]{babel}

\usepackage{amsthm}

\DeclarePairedDelimiter\floor{\lfloor}{\rfloor}

\usepackage{subcaption}

\newcommand{\no}{\noindent}

\newtheorem{probform}{Formulation}[section]


%% file: 1_Introduction.tex
\section{Introduction} \label{sec:intro}

Mobile robots have been used to navigate in indoor environments (such as households, offices, hospitals, etc. \cite{zhang2020wifi,ramdani2019safe,conte2020design}), and outdoor environments such as agricultural fields, forests, etc. \cite{r2018research,champ2020instance,couceiro2019semfire}. Such complex environments contain obstacles of various sizes, densities/solidities, and opacities that are challenging in terms of the robot's perception, and navigation. For instance, contemporary indoor environments contain objects such as string/beaded curtains, transparent surfaces such as glass walls \cite{mei2020glass_vision,kuribayashi2022corridor}, etc. Outdoor scenarios, on the other hand, have complex vegetation such as pliable tall grass, bushes, trees, etc. in close proximity to and intertwined with each other. A major challenge, and an important requirement for autonomous navigation, is differentiating \textit{truly solid} and impassable obstacles (furniture, glass surfaces, bushes, trees, etc.) from obstacles that can be passed through (beaded curtains, tall grass, etc.).

To first detect obstacles, mobile robots have predominantly used RGB and depth images \cite{hua2019small}, 2D lidar scans \cite{fox1997dwa}, 3D point clouds \cite{wu2020deep3dlidar_object}, etc. The raw data from these sensors has been used to: (1). Estimating the proximity of objects in the robot's vicinity; (2). Computing a variation of an occupancy grid \cite{gerkey2008planning_dwa} or a cost map representation that indicates both an obstacle's size and distance from the robot; or (3). Segmenting the scene to assess obstacles' size, distance, and semantic meaning \cite{ying2022uctnet_segmentation,guan2022ga}. Although such approaches have been used to aid navigation, they may not work well in environments composed of thin, pliable/bendable, and transparent obstacles. For instance, time-of-flight sensors such as depth cameras and lidars tend to detect thin, pliable, and passable obstacles (e.g. string curtains, tall grass) as solid obstacles that the robot must avoid \cite{petty2022lidar}. On the other hand, transparent objects such as glass remain undetected since the laser rays mostly pass through them, leading to collisions during navigation. Similarly, perception methods using RGB images may not work well in terms of detecting transparent objects and visually similar but structurally dissimilar vegetation. Moreover, they are severely affected by the environment's lighting changes.

\begin{figure}[t]
    \centering
    \includegraphics[width=\columnwidth,height=5.5cm]{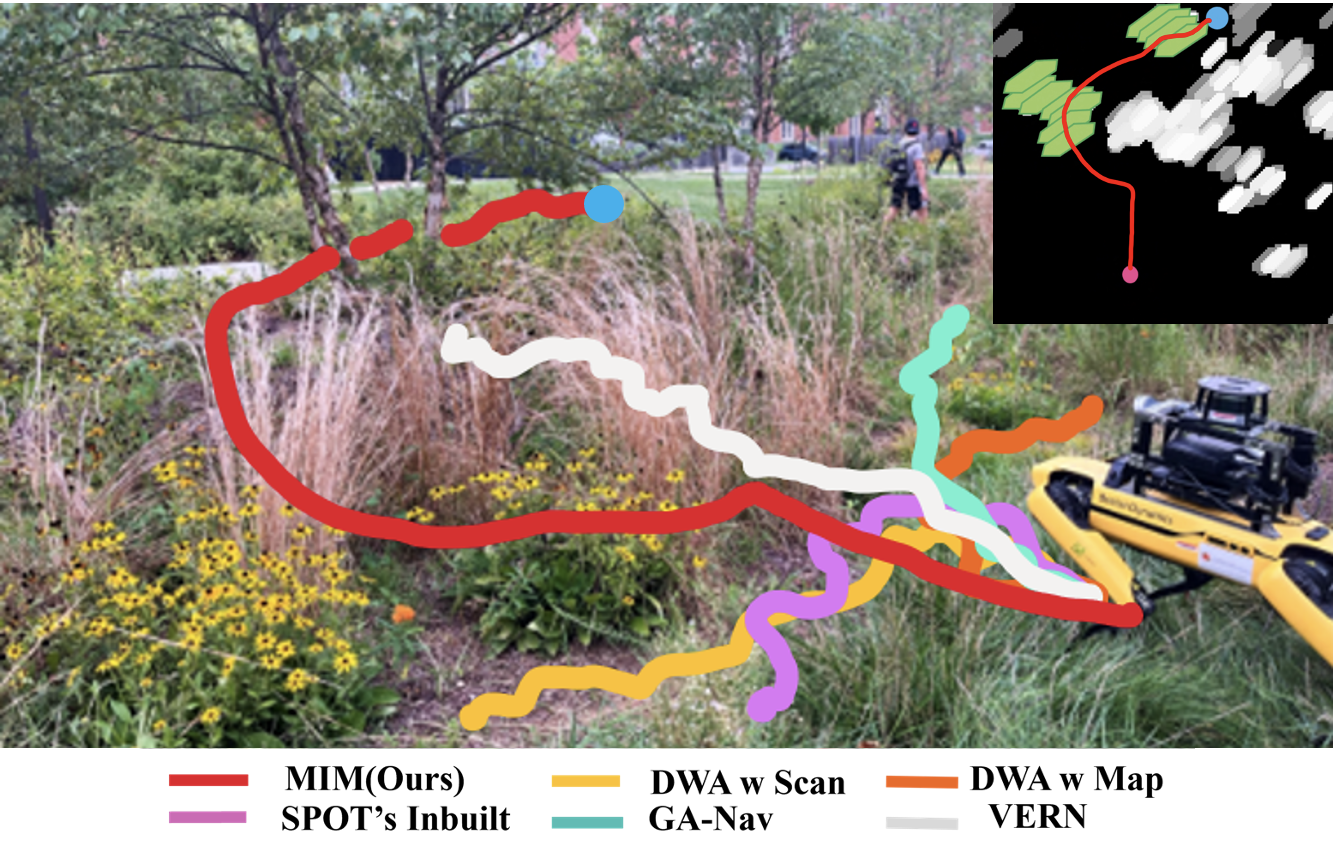}
    \caption{\small{Comparison of trajectories while navigating using our Multi-layer Intensity Maps, DWA with laser scan \cite{fox1997dwa}, DWA with occupancy map \cite{gerkey2008planning_dwa}, Spot's Inbuilt Autonomy, and VERN \cite{jagan2023vern} in complex vegetation. The intensity map for the scenario is shown in the top right, with the robot in pink, its goal in blue, passable obstacles such as tall grass in green, and the robot's trajectory overlayed for reference. intensity maps help differentiate solid objects such as trees even when they are intertwined with tall grass. Other methods for robot perception in this case either freeze or collide with the solid obstacles.}}
    \label{fig:cover_image}
\end{figure}

\textbf{Main Contributions:} In order to address these robot perception challenges, we present a novel obstacle representation called \textit{multi-layer intensity map} that can be used to simultaneously estimate the size/height, density/solidity, and opacity of objects in the environment. The intensity maps are constructed by stacking individual layers of 2D grid maps, each computed from reflected point clouds intensities corresponding to a certain height interval. It preserves the benefits of existing occupancy grids such as indicating the size and proximity of obstacles around the robot while also accurately detecting passable obstacles as such. The novel components of our approach include:

\begin{itemize}
    \item A novel obstacle representation called the multi-layer intensity map constructed from the intensities of the reflected points in a point cloud. In addition to an object's height and size, the multi-layer intensity map also reflects its true density/solidity. Our multi-layer intensity map can replace occupancy grids and other map representations in existing navigation methods to enable a robot to navigate through passable/navigable indoor and outdoor obstacles that are often misclassified by existing representations.

    \item A novel method to detect transparent objects using the low-intensity reflected points in the multi-layer intensity map. Our approach accurately extrapolates the transparent surfaces from a small neighborhood of low-intensity points which enables a robot's motion planner to avoid collisions, significantly improving its rate of safely reaching its goal.

    \item A novel method using the multi-layer intensity map to accurately identify objects that are safe to navigate through such as thin, passable curtains in indoor scenarios, and pliable tall grass in outdoor environments. Our approach alleviates robot freezing behaviors in the presence of such objects.

    \item An adaptive inflation strategy that assesses the detected solid obstacles (e.g. concrete and glass walls, furniture indoors, and bushes and trees outdoors) to enlarge them in the multi-layer intensity map for efficient planning. Our strategy handles narrow scenarios such as doors, corridors, and passages between dense vegetation and trees, where existing navigation schemes freeze. We demonstrate significant improvements in navigation using intensity maps in indoor scenes using a real Turtlebot, and in complex outdoor environments using a Boston Dynamics Spot robot.
\end{itemize}

%% file: 2_Related_Work.tex
\section{Related Work}
In this section, we give a brief overview of perception methods that use point cloud intensities. Additionally, we review the existing obstacle detection research in indoor and outdoor settings. 

\subsection{Sensing using Point Cloud Intensity}


The importance of intensity information in LiDAR point cloud data has been an emerging focus in robotics \cite{di2021visual,sullivan2014fusing,laible20123d}. Some methods investigate the use of intensity information alongside geometric features to enhance point cloud classification methods in outdoor settings \cite{reymann2015improving}. These methods highlight the potential of using the intensity information to provide a better understanding of obstacles, particularly when scene illumination is not consistent. Other methods include the ISHOT descriptor \cite{guo2019local}, which combines geometric and intensity data for improved place recognition.
Lidar intensity maps have been also used for localization \cite{9129063,wei2017estimating,barsan2020learning}. In \cite{9129063}, the authors present a robust Graph-SLAM framework that improves map accuracy for autonomous vehicles by encoding road surfaces based on LIDAR reflectivity. Moreover, the application of LiDAR intensity in visual navigation tasks has been explored. For instance, \cite{barfoot2016into} introduces a lidar-intensity-image pipeline and demonstrates its performance in visual odometry (VO) and visual teach and repeat (VT\&R) tasks. Lidar intensity maps have also been leveraged for various other applications, including orthoimage generation \cite{shin2021true} and anisotropic surface detection \cite{garestier2014anisotropic}.


\subsection{Detecting Indoor Obstacles} 

Object detection in indoor settings has been widely studied for numerous applications including robot navigation, mapping, and computer graphics. Popular solutions in the literature include vision-based object detection and semantic segmentation approaches due to the structuredness of indoor environments. Moreover, the generation of necessary image datasets is feasible due to the limited diversity of indoor objects. However, detecting non-opaque objects such as glass remains a formidable challenge for vision-based systems due to the lack of visual clues. The method in \cite{mei2020glass_vision} proposes GDNet, a glass detection network that identifies abundant contextual cues for glass detection using a large-field contextual feature integration (LCFI) module. UCTNet \cite{ying2022uctnet_segmentation} proposes a  cross-modal transformer network for indoor RGB-D semantic segmentation to identify different objects such as curtains, doors, etc. However, such methods require large datasets with pixel-level ground truth labeling.

\subsection{Detecting Outdoor Obstacles}

Over recent years, there has been significant progress in the  development of robotic systems designed for outdoor navigation \cite{guan2022ga, dupeyroux2019ant,li2021openstreetmap,huskic2019gerona,weerakoon2022graspe}. One early approach can be found at \cite{wurm2009improving}, where the usage of laser measurements enabled navigation capabilities for robots such as detecting short and grass-like vegetation. However, this method is not universally applicable, particularly in complex, unstructured vegetative terrains. 
Complementary approaches have tackled associated issues in off-road navigation, specifically concerning varying slopes \cite{terp} and different terrains \cite{terrapn}.

Many studies integrate proprioceptive with exteroceptive sensory data to enhance outdoor navigation \cite{robust-perceptive-locomotion-quad, homberger2019support, elnoor2023pronav}. Machine learning techniques have also been incorporated to augment the robot's capabilities for navigating through pliable vegetative obstacles \cite{badgr,model-error-katyal}. 
To this end, \cite{jagan2023vern} uses a few shot learning approach to classify RGB vegetation images based on their traversability. This classifier is then integrated with a 3D LiDAR to construct a vegetation-aware traversability cost map. 


%% file: 3_Background.tex
\section{Background}

\subsection{Definitions and Assumptions}
Our formulation assumes that a sensor capable of generating 3D point clouds (e.g., 3D lidar, depth camera) is mounted on a robot with a 2D linear $(v)$ and angular $(\omega)$ velocity space. Rigid coordinate frames are attached to the robot and sensor with the positive $x, y, z$ directions facing forward, left, and upwards respectively, and for simplicity, we assume both frames to coincide. All positions, and velocities are measured relative to these frames. At any time $t$, the robot has a preferred velocity direction aimed at its goal $(g_x, g_y)$ as $\theta = \tan^{-1}(g_y/g_x)$. 

Our approach is based on using the reflected intensities of point clouds that could be obtained from sensors such as 3D lidars, depth cameras, etc that have a laser source/transmitter and a receiver. We represent a point $\mathbf{p}$ in a point cloud as $\mathbf{p} = \{x, y, z, int\}$, where $x, y, z$ denote the point's location relative to the sensor, and $int \in [0, R] $ denotes its intensity. We define a 2D robot-centric grid map as containing $n \times n$ grids. Each grid is denoted by a row $r$ and column $c$. Each grid represents a $g \times g$ area in the real-world, and the value contained in it indicates the probability of the presence of an obstacle. Finally, we use $j$ and $k$ to denote indices.

\subsection{Point Cloud Intensity}
Typically, a point's intensity is high ($int > 0.75R$) when it is reflected from solid, opaque, 3D (length, width, and height dimensions are not infinitesimal) objects since they prevent the sensor's laser rays from passing through, or scattering away from the sensor. In contrast, objects that are low density (e.g. tall grass which is a collection of thin blades of grass that scatter laser rays), and transparent (e.g. glass where laser rays mostly pass through) lead to low intensities ($int < 0.5R$) or in some cases, no intensity ($int = 0$).

\subsection{Obstacle Properties}
Our formulation leverages the property to accurately detect truly solid objects from the following categories defined based on how objects are sensed by existing perception modalities (2D lidar scan, RGB and depth images, etc) as:

\begin{itemize}
    \item \textbf{True Positives (TP)}: Solid, non-traversable objects detected as solid, e.g. walls, wooden furniture, etc.
    \item \textbf{True Negatives (TN)}: Non-solid, traversable objects detected as passable or no obstacle, e.g. free space.
    \item \textbf{False Positives (FP)}: Non-solid objects detected as solid, e.g. string/beaded curtains, pliable tall grass.
    \item \textbf{False Negatives (FN)}: Solid objects detected as non-solid or as free space, e.g. transparent objects.
\end{itemize}

\subsection{Obstacle Inflation} \label{sec:inflation-bg}
Once a truly solid object is detected, it must be enlarged or \textit{inflated} for the robot's planner to ensure that it avoids it at a safe distance \cite{inflation}. Inflation is performed prior to planning to expand obstacles uniformly in all dimensions by a certain amount (typically the robot's radius, or maximum(length, width) to ensure that the planner avoids obstacles by a safe distance. Standard methods for obstacle inflation include performing the Minkowski sum \cite{minkowski-sum} between the robot's radius and the obstacle, cost propagation from the obstacle, dilating obstacles using convolutions, etc. on a grid map. With these preliminaries, we state our problem formulation as follows:  

\begin{probform}
To construct an $n \times n \times m$ grid map representation $I^t_{ML}$ of obstacles from points $\mathbf{p} = \{x, y, z, int\}$ and classify each grid $(r, c) \in I^t_{ML}$ as a true positive ($TP$), false positive ($FP$), false negative ($FN$) obstacle, or true negative ($TN$) free space and enlarge $TP$ and $FN$ obstacles adaptively based on the robot's preferred velocity direction.
\end{probform}

%% file: 4_Intensity_Map.tex
\section{Our Approach}
\begin{figure}[t]
    \centering
    \includegraphics[width=\columnwidth,height=3.60cm]{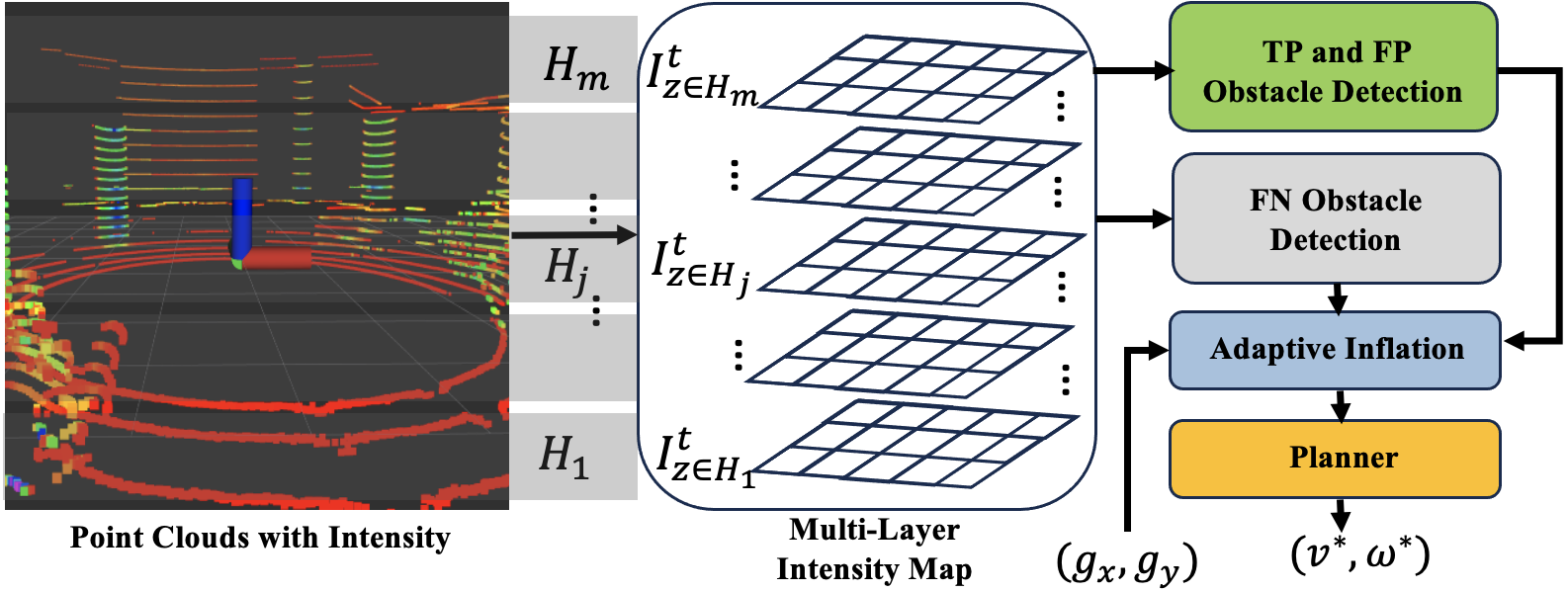}
    \caption{\small{Our approach's overall system architecture. At time instant $t$, the intensities of reflected point clouds from a height interval $H_j$ (grey rectangles) are used to construct a 2D grid map layer $I^t_{z \in H_j}$ according to equation \ref{eqn:single-layer}. Several of such layers are stacked to form an intensity map. Certain layers from intensity maps can be used to detect TP, FP, and FN obstacles. The adaptive inflation enlarges the truly solid (TP, FN) obstacles that the robot must avoid in the direction opposite to the robot's goal $(g_x, g_y)$` direction, and ignores passable obstacles (FP). The planner finally uses the inflated map to compute collision-free linear and angular velocities $v^*, \omega^*$.}}
    \label{fig:sys-arch}
\end{figure}

In this section, we discuss how 3D point cloud intensities can be used to construct the multiple 2D grid map layers of an intensity map. The input point clouds could be obtained from any sensor such as a 3D lidar or a depth camera that also measures the intensity of the points reflected from surrounding objects. We show how different layers of the intensity map can be used to accurately differentiate solid obstacles from passable objects, and identify transparent obstacles. Finally, we detail our obstacle inflation strategy, which enables robots to navigate through passable obstacles, and narrow passages. Fig. \ref{fig:sys-arch} shows our overall proposed architecture.

\subsection{Multi-Layer Intensity Map}
We construct a single layer of the intensity map corresponding to a height interval $H_j$ at any time instant $t$ as follows,
\begin{equation}
\begin{split}
    I^t_{z \in H_j}(r, c) &= \frac{\sum_{x}\sum_{y} int}{g^2} \\ 
    \forall \,\, x &\in [x_{low}, x_{low} + g], \,\, y \in [y_{low}, y_{low}+g], \\ 
    x_{low} &= \floor*{(r - \frac{n}{2}) \cdot g} \,\,\, \text{and} \,\,\, y_{low} = \floor*{(c - \frac{n}{2}) \cdot g} \\
    H_j &= [low_j, high_j], \,\, low_j \le high_j,
\end{split}
\label{eqn:single-layer}
\vspace{-15pt}
\end{equation}
\no where $\floor*{}$ represents the floor operation, $low_j$, and $high_j$ are the limits for all the points whose intensities must be considered for the map along the $z$ direction. Extending this definition, we construct a \textit{multi-layer} intensity map as a stacking of $m$ layers as,

\begin{equation}
    I^t_{ML}(r, c) = [I^t_{z \in H_1}(r, c) \,|\, ... \,|\, I^t_{z \in H_j}(r, c) \, | \, ... \, | \, I^t_{z \in H_m}(r, c)],
    \label{eqn:multi-intense}
\end{equation}

\no where $H_1,..., H_j, ..., H_m$ are non-overlapping height intervals. We choose stack multiple non-overlapping layers at various heights instead of combining the points' intensities at all heights into a unified layer. This is due to the flexibility that multiple layers provide in analyzing and modifying them individually. Furthermore, individual layers can be combined after modification and used for planning the robot's trajectories. We highlight these benefits in the following sections.

\subsection{Obstacle Detection}
In this section, we describe how challenging obstacles such as tall grass (FP), string/beaded curtains (FP), and transparent objects (FN) can be detected using our multi-layer intensity map.

\subsubsection{Differentiating True and False Positive Obstacles} 
Existing methods that use various sensor modalities typically detect many thin, pliable obstacles such as tall grass, and passable objects such as string/beaded curtains as solid obstacles that must be avoided. During navigation, such inaccurate detections cause the robot to freeze or oscillate perpetually without reaching its goal. We show how such false positive obstacles can be detected using the multi-layer intensity map. 

Let us consider three layers $I^t_{z = 0}, I^t_{z \in (0, h]}, I^t_{z \in [-h, 0)}$ of the intensity map. If a grid location $(r, c)$ belonging to all three layers satisfies the following condition $\mathcal{C}$, we classify that grid as a false positive obstacle:

\begin{equation}
  \mathcal{C}(r, c): I^t_{z = 0}(r, c), \,\, I^t_{z \in (0, h]}(r, c), \,\, I^t_{z \in [-h, 0)}(r, c) \le \Gamma.
\end{equation}

\no Here, $\Gamma$ is an intensity threshold. Using this condition, we construct TP and FP intensity maps for planning as,

\begin{equation}
\begin{split}
    I^t_{TP}(r, c) = max(I^t_{z = 0}(), I^t_{z \in (0, h]}(), I^t_{z \in [-h, 0)}()) \\ \forall \,\, (r, c) \,\, | \,\, \mathcal{C}(r, c) \,\, \text{is False}
\end{split}
\end{equation}

\begin{equation}
\begin{split}
    I^t_{FP}(r, c) = max(I^t_{z = 0}(), I^t_{z \in (0, h]}(), I^t_{z \in [-h, 0)}()) \\ \forall \,\, (r, c) \,\, | \,\, \mathcal{C}(r, c) \,\, \text{is True}
\end{split}
\end{equation}

\subsubsection{Detecting False Negative Obstacles} \label{sec:glass-detection}
False negative obstacles are typically transparent objects that allow most of the laser rays from a sensor to pass through. However, a small neighborhood of points with low intensities are detected for laser rays that are incident $\sim 0^{\degree}$ on a transparent surface only along the $z = 0$ plane \cite{lasitha2022cartographer_glass}. Our approach extrapolates this small neighborhood to detect solid transparent obstacles.

Consider two layers $I^t_{z=0}$ and $I^t_{z=-\epsilon}$ in the multi-layer intensity map at time instant $t$. Here, $\epsilon$ is a small positive value. To isolate the low-intensity neighborhood of points reflected from the transparent object, we first calculate the element-wise difference between the two layers $I^t_{z=0} \ominus I^t_{z=-\epsilon}$. This removes all the points reflected from the same obstacles in both the layers and retains only the points corresponding to the small glass neighborhood. To indicate the presence of transparent obstacles for subsequent time steps, we transform $I^t_{z=0} \ominus I^t_{z=-\epsilon}$ based on the robot's motion as,

\begin{equation}
    I^t_{FN} = T \cdot (I^t_{z=0} \ominus I^t_{z=-\epsilon}), 
\end{equation}

\no where T is a $4 \times 4$ transformation matrix whose rotational component is based on the robot's yaw, the translational component is based on its motion from time $t$ to $t + 1$. Finally, we augment subsequent time's $I^{t+1}_{FN}$ using $I^t_{FN}$ as,

\begin{equation}
    I^{t+1}_{FN} = I^{t+1}_{FN} \,\, \bigcup \,\, I^{t}_{FN}.
\end{equation}


\subsection{Adaptive Obstacle Inflation}
\begin{figure}[t]
    \centering
    \includegraphics[width=\columnwidth,height=2.5cm]{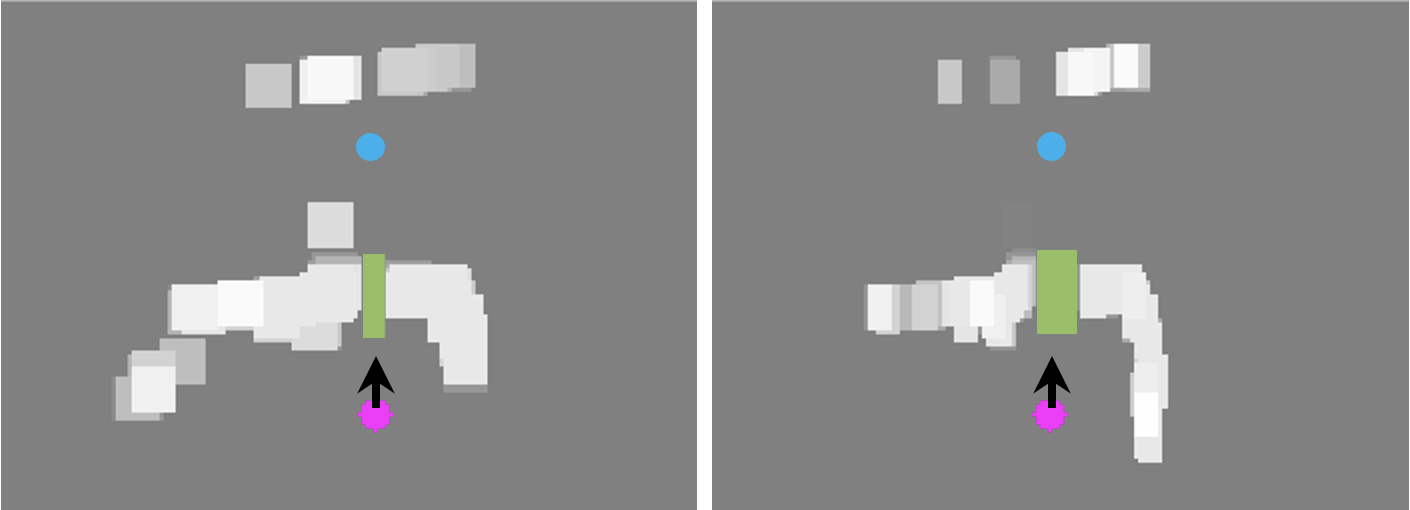}
    \caption{\small{The robot's position and its goal are denoted by the pink and blue circles respectively. The robot's heading and goal direction coincide in this case. \textbf{[Left]}: Uniformly inflating obstacles using an $e \times e$ kernel with all ones near a narrow passage closing up the available free space (green rectangle). \textbf{[Right]}: Adaptively inflating the obstacles based on the robot's goal direction preserves the free space while inflating the obstacles in the direction opposite to the robot's goal direction.}}
    \label{fig:inflation}
\end{figure}

In narrow and dense scenarios, uniformly inflating obstacles (as explained in section \ref{sec:inflation-bg}) could close the available free space (see fig. \ref{fig:inflation}) leading to the robot freezing problem \cite{frozone}. If obstacles are not inflated, the robot could move close to the obstacles and even collide with them. Additionally, false positive obstacles that can be traversed need not be inflated. Therefore, our approach adaptively inflates the obstacles based on the robot's goal direction. Obstacles are majorly inflated in the direction opposite to the robot's goal/preferred direction, and minorly inflated in all other directions. This ensures that the robot does not navigate too close to a solid obstacle, while also never closing free space near the narrow passages. 

Our inflation is performed using the convolution operation on obstacles by computing the appropriate kernel matrices ${K}$ of size $e \times e$ as follows. Let $(g_x, g_y)$ be the goal location relative to the robot's coordinate frame. The goal direction can be defined by the slope $\tan{\frac{g_y}{g_x}}$. To design a kernel matrix ${K}$ to inflate obstacles in the opposite direction, we first find the line along the goal direction relative to the kernel, passing through its center $(\frac{e}{2}, \frac{e}{2})$ as, 

\begin{equation}
\begin{split}
    f(r^{{K}}, c^{{K}}) &:  c^{{K}} - \tan(g_y/g_x) \cdot r^{{K}} + \text{const} = 0 \\
    \text{const} &= \frac{e}{2}(\tan(g_y/g_x) - 1).
\end{split}
\end{equation}

\no Here, $(r^K, c^K)$ represent the row and column on the kernel. Next, the kernel can be constructed as,

\begin{equation}
    {K}(r^{{K}}, c^{{K}}) = 
    \begin{cases}
    1 \quad \forall \{r^{{K}}, c^{{K}} | f(r^{{K}}+j, c^{{K}}+j) = 0\} \\
    0 \quad \text{Otherwise}.
    \end{cases}
\end{equation}

\no Here, $j \in [\text{-padding}, \text{padding}]$ is added to the row and column indices to control the level of thickness to inflate an obstacle. Finally, using ${K}$, we convolve our multi-layer intensity map as,

\begin{equation}
    I^t_{inflate} = (I^t_{TP} \bigcup I^t_{FN}) \circledast K.
\end{equation}

\no $I^t_{inflate}$ contains inflated True Positive and False Negative obstacles. To add information about false positive obstacles prior to planning, we perform,

\begin{equation}
    I^t_{plan} = I^t_{inflate} \,\, \bigcup \,\, I^t_{FP},
\end{equation}

\no where $I^t_{plan}$ is a 2D grid map containing TP, FP, FN, and free space represented by various grids, which can be used as a cost map for motion planning.

\begin{figure*}[t]
    \centering
    \includegraphics[width=1.5\columnwidth,height=6cm]{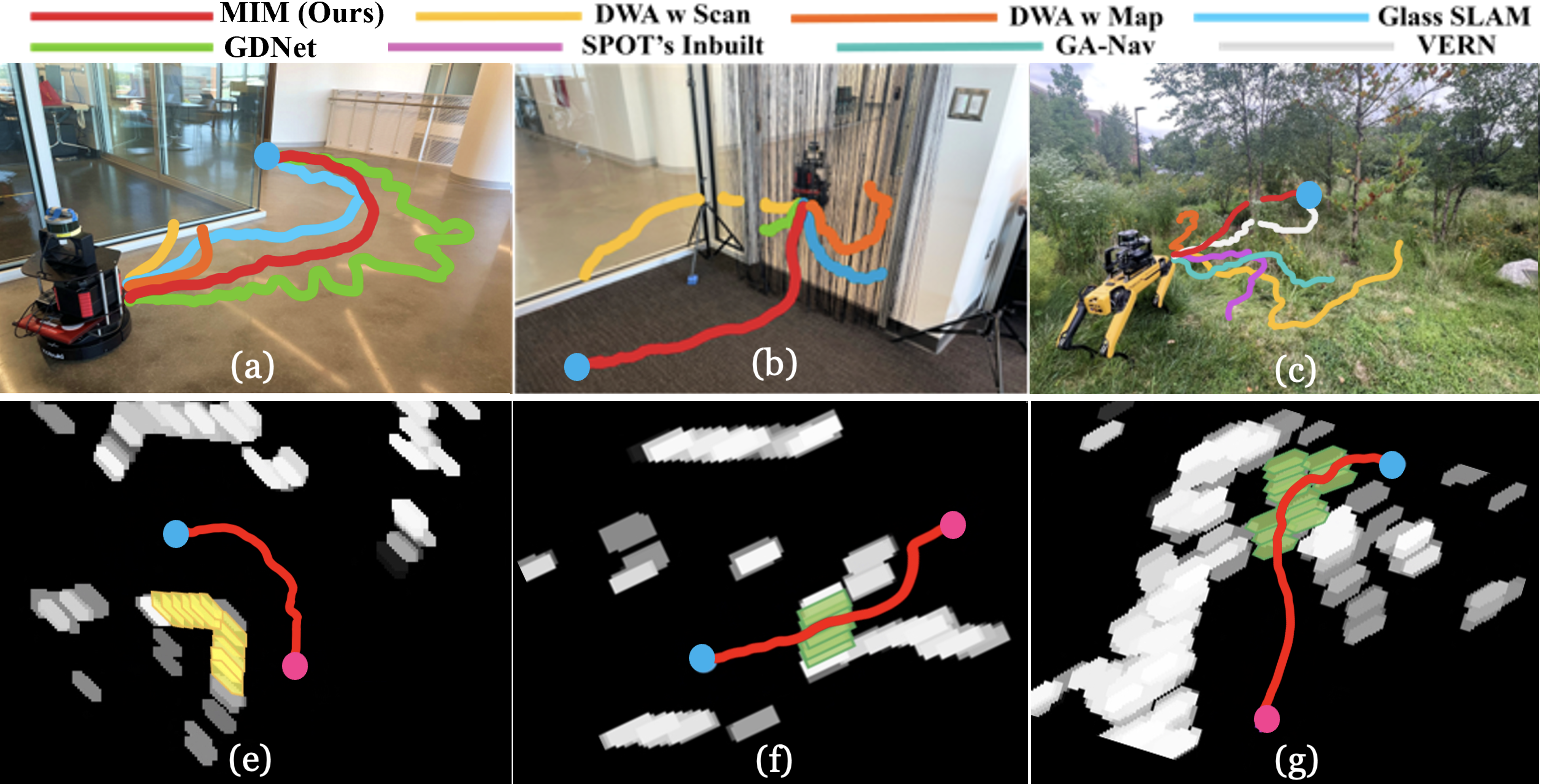}
    \caption{\small{Robot trajectories when navigating in different complex indoor and outdoor environments using various methods. (a) Intensity map identifies transparent objects such as glass in real-time and avoids it, (b) Intensity map identifies string curtains as safe, passable obstacle while other methods detect it as solid, and impassable. (c) Intensity map accurately identifies pliable tall grass regions (which registers lower intensities on $I^t_{plan}$) to navigate through, avoiding trees. (d,e,f) show the corresponding $I^t_{plan}$ for the scenarios above. White, grey, and black colors represent intensities in a decreasing order of magnitude. The robot's starting location is in pink, and goal is represented in blue. The yellow in (d) represents the glass (FN obstacle) extrapolated in real-time by our method in section \ref{sec:glass-detection}. Green in (e, f) represent passable, non-solid obstacles such as curtains and tall grass respectively. Intensity map-based navigation's trajectory is overlayed for reference. (d, e, f) also depict obstacles inflated based on the robot's goal direction.}}
    \label{fig:comparison_trajs}
\end{figure*}

\subsection{Integration with Planning Methods} \label{sec:planner-integration}
The final planning intensity map $I^t_{plan}$ can be used with any motion planner as a cost map to evaluate a candidate linear and angular velocity's $(v, \omega)$ obstacle or collision cost. This can be computed by extrapolating the trajectory produced by $(v, \omega)$ relative to $I^t_{plan}$ as in \cite{terrapn,jagan2023vern} as, 

\begin{equation}
    \begin{split}
        traj^{I^t_{plan}} = [(r_1, c_1), ... , (r_j, c_j), ... , (r_p, c_p)] \\
        \text{Obstacle Cost} = \sum_{j=1}^{p} I^t_{plan}(r_j, c_j).
    \end{split}
\end{equation}

Additionally, $I^t_{plan}$ can be used as an observation with deep reinforcement learning-based navigation methods \cite{terp} to improve the trained model's understanding of the solidity of surrounding obstacles.

%% file: 5_Results.tex
\section{Results and Analysis}
In this section, we summarize our method's implementation details and evaluation metrics. Then, we present the details of the experiments conducted to highlight the benefits of our approach.

\subsection{Implementation} \label{sec:implementation}
We use a Boston Dynamics Spot robot for outdoor experiments and a Turtlebot 2 robot for indoor experiments. The Spot robot is equipped with an Intel NUC 11 with Intel i7 CPU and NVIDIA RTX 2060 GPU. The Turtlebot is equipped with a laptop with an Intel i9 CPU and an Nvidia RTX 2080 GPU. Both robots are equipped and a Velodyne VLP16 lidar. We particularly use Velodyne lidar's dual return mode since it captures low intensities in the point cloud better \cite{lasitha2022cartographer_glass}. We use $n = 200$, and $m = 4$ layers in our implementation (one for $I^t_{z \in [-h, 0)}, I^t_{z \in (0, h]}, I^t_{z = 0}, I^t_{z = -\epsilon}$). The height ranges are adjusted based on the robot's height.

\subsection{Comparison Methods and Evaluation Metrics}
We compare the improvements in navigation while using intensity maps with various existing methods. We use intensity maps with the dynamic window approach (DWA) \cite{fox1997dwa} and calculate candidate velocity's obstacle costs as described in section \ref{sec:planner-integration}. For fair comparison, we use DWA as the baseline planner and integrate other methods for obstacle perception with it for navigation. We compare with various indoor and outdoor navigation methods that use a variety of perception inputs such as RGB/depth image, 2D lase scans, or occupancy grids created by any proximity sensor.


DWA is a local planner that can use a 2D LiDAR scan \cite{fox1997dwa} or an occupancy map \cite{gerkey2008planning_dwa} to perform obstacle avoidance. Spot's in-built autonomy incorporates a set of stereo cameras around the robot to estimate the obstacles and ground plane to navigate to a goal. VERN is a vegetation-aware navigation approach that uses an RGB image-based vegetation classifier and a set of 2D occupancy grid maps for perception. In indoor scenarios, Glass-SLAM \cite{WANG201797_glass_detection_laser_intensity} uses the specular reflection of laser beams from the glass to map environments that include glass. GDNet \cite{gdnet} is a semantic segmentation method that uses RGB images to segment glass from a scene. We also perform ablations studies between using adaptive and uniform inflation using intensity maps. 

\no \textbf{Success Rate} - Number of successful goal-reaching attempts (without collisions with solid objects or freezing behaviors) out of the total number of trials.

\no \textbf{Normalized Traj. Length} - The ratio between the robot's trajectory length and the straight-line distance to the goal from the starting location averaged over all runs.  

\no \textbf{F-Score} - A measure of object detection accuracy calculated as a weighted average of the precision and recall. Values are between 0 and 1, where 1 indicates the best accuracy. We use human detection of TP, FP and FN obstacles to calculate precision and recall.

\no \textbf{Inference Time} - Time taken from the instant an obstacle is viewed to the instant when it is detected.

\subsection{Testing Scenarios}

\begin{itemize}
\item \textbf{Scenario 1} - Indoor scenario with glass, concrete walls, and pillars (Fig. \ref{fig:comparison_trajs}a).

\item \textbf{Scenario 2} - Indoor scenario with a narrow passage covered with a beaded/string curtain (Fig. \ref{fig:comparison_trajs}b).

\item \textbf{Scenario 3} - Outdoor scenario with tall grass, bushes, and trees separated from each other (Fig. \ref{fig:comparison_trajs}c). 

\item \textbf{Scenario 4} - Outdoor scenario with tall grass, bushes, and trees closely intertwined with each other creating narrow passages (Fig. \ref{fig:cover_image}). 
\end{itemize}

\begin{table}
\resizebox{\columnwidth}{!}{%
\begin{tabular}{ |c |c |c |c |c |} 
\hline
\textbf{Scenario} & \textbf{Method} & \multicolumn{1}{|p{1.5cm}|}{\centering \textbf{Sucess} \\ \textbf{Rate (\%)}} & \multicolumn{1}{|p{1.5cm}|}{\centering \textbf{Norm. Traj.} \\ \textbf{Length}} & \multicolumn{1}{|p{1cm}|}{\centering \textbf{F-Score} } 
\\ [0.5ex] 
\hline
\multirow{7}{*}{\rotatebox[origin=c]{0}{\makecell{\textbf{Scn.}\\\textbf{1}}}} 
 & DWA with laser scan \cite{fox1997dwa} & 0 & 0.342 & 0.12   \\
 & DWA with occupancy map \cite{gerkey2008planning_dwa} & 0 & 0.415 & 0.20 \\
 & Glass-SLAM \cite{WANG201797_glass_detection_laser_intensity} & 30 & 0.895 & 0.76  \\
 & GDNet \cite{gdnet} & 50 & 0.633 & 0.69 \\
 & Intensity map (Uniform) & \textbf{80} & 1.256 & \textbf{0.82}  \\
 & Intensity map (Adaptive) & \textbf{80} & \textbf{1.052} & \textbf{0.82}  \\
\hline

\multirow{7}{*}{\rotatebox[origin=c]{0}{\makecell{\textbf{Scn.}\\\textbf{2}}}} 
 & DWA with laser scan \cite{fox1997dwa} & 0 & 0.765 & 0.17   \\
 & DWA with occupancy map \cite{gerkey2008planning_dwa} & 0 & 0.355 & 0.16 \\
 & Glass-SLAM \cite{WANG201797_glass_detection_laser_intensity} & 0 & 0.421 & 0.46  \\
 & GDNet \cite{gdnet} & 0 & 0.318 & 0.65 \\
 & Intensity map (Uniform) & 30 & \textbf{1.134} & \textbf{0.79}  \\
 & Intensity map (Adaptive) & \textbf{70} & 1.141 & \textbf{0.79}  \\
\hline

\multirow{7}{*}{\rotatebox[origin=c]{0}{\makecell{\textbf{Scn.}\\\textbf{3}}}} 
 & DWA with laser scan \cite{fox1997dwa} & 20 & 1.571 & 0.35   \\
 & DWA with occupancy map \cite{gerkey2008planning_dwa} & 20 & 1.482 & 0.33 \\
 & Spot's Inbuilt Autonomy  & 10 & 0.311 & 0.38  \\
 & VERN \cite{jagan2023vern} & 70 & 1.154 & 0.75 \\
 & Intensity map (Uniform) & 40 & 1.292 & \textbf{0.84} \\
 & Intensity map (Adaptive) & \textbf{80} & \textbf{1.072} & \textbf{0.84}  \\
\hline

\multirow{7}{*}{\rotatebox[origin=c]{0}{\makecell{\textbf{Scn.}\\\textbf{4}}}} 
 & DWA with laser scan \cite{fox1997dwa} & 0 & 1.543 & 0.37   \\
 & DWA with occupancy map \cite{gerkey2008planning_dwa} & 0 & 1.412 & 0.36 \\
 & Spot's Inbuilt Autonomy & 0 & 0.298 & 0.31  \\
 & VERN \cite{jagan2023vern} & 50 & 1.267 & 0.68 \\
 & Intensity map (Uniform) & 60 & 1.248 & \textbf{0.74}  \\
 & Intensity map (Adaptive) & \textbf{70} & \textbf{1.146} & \textbf{0.74}  \\
\hline
\end{tabular}
}
\caption{\small{Performance comparison between using intensity maps for navigation versus other methods in indoor and outdoor scenarios using various metrics. We observe that intensity maps are versatile in detecting a variety of perceptionally challenging obstacles, and aiding the navigation.}
}
\label{tab:comparison_table}
\end{table}

\subsection{Analysis and Comparison} \label{sec:analysis}

We present the qualitative navigation experiment results for the four scenarios in Fig. \ref{fig:comparison_trajs} and the quantitative results in Table \ref{tab:comparison_table}. Scenarios 1 and 2 are complex indoor settings, whereas scenarios 3 and 4 are outdoor settings. We observe that intensity map demonstrates the highest success rate compared to the other methods in all four scenarios.

\begin{figure*}[t]
    \centering
    \includegraphics[width=\linewidth,height=6cm]{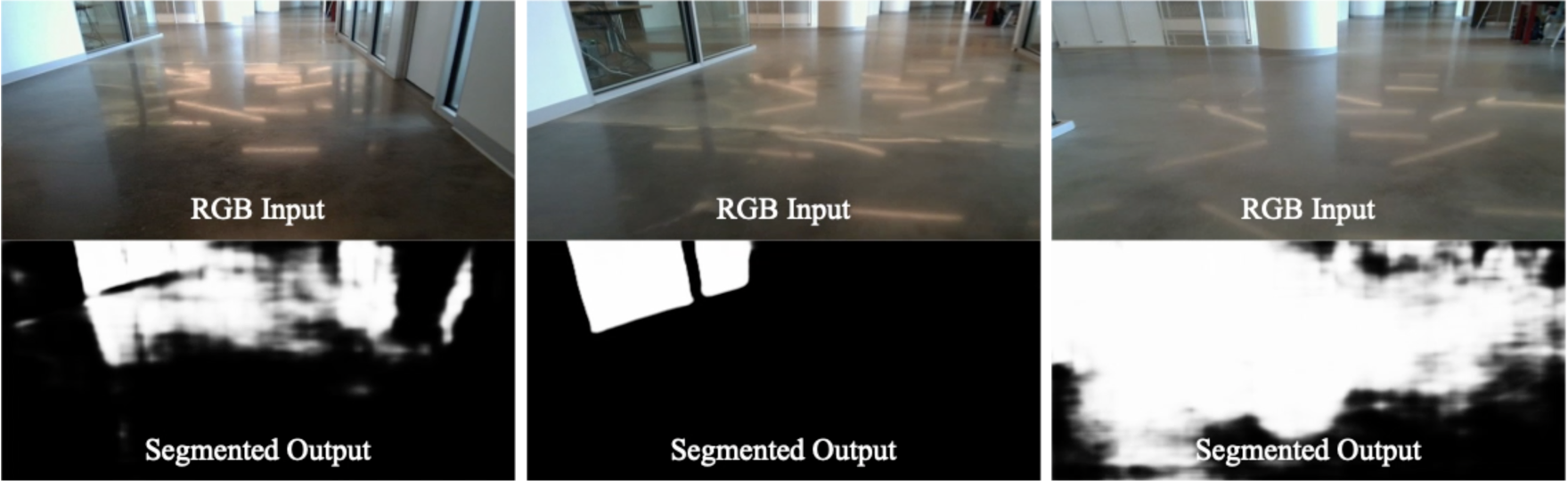}
    \caption{\small{The results of GDNet \cite{gdnet} in various instances in scenario 1. While GDNet accurately segments glass in some instances ([middle]), it is often inaccurate due to reflections from the floor ([left], [right]).}}
    \label{fig:gdnet-results}
\end{figure*}

In scenario 1, 2D laser scan-based and occupancy map-based DWA planners fail to identify the glass region since they do not incorporate lidar intensity for object detection. GDNet based indoor segmentation fails to identify the glass region from RGB inputs consistently due to lighting changes, and strong lights reflected from the floor (see Fig. \ref{fig:gdnet-results}). Hence, these methods lead to collisions with the glass by assuming it to be free space. Glass-SLAM can identify the glass region using lidar intensity. However, it does not avoid glass all the time since the glass is constructed slower than the robot's motion (see Fig. \ref{fig:glass-slam-results}). Intensity map's multi-layer map formulation leads to consistent glass detection which also results in a higher F-score compared to the other methods. 

\begin{table}[]
\centering
\resizebox{0.75\columnwidth}{!}{
\begin{tabular}{|c|c|c|}
\hline
\textbf{Method} & \textbf{Inference Time (sec)} & \textbf{Training Time}  \\
\hline
Intensity map &  0.020 on CPU & NA\\
\hline
Glass-SLAM & $4.5$ on CPU & NA\\
\hline
GDNet & 0.112 on GPU & 12-15 hours \\
\hline
VERN &  0.083 on GPU & 6-8 hours\\
\hline
\end{tabular}
}
\caption{\small{The inference rates and training time (where applicable) for several methods that detect glass (Glass-SLAM \cite{WANG201797_glass_detection_laser_intensity}, GDNet \cite{gdnet}), and pliable vegetation (VERN \cite{jagan2023vern}). intensity maps are capable of detecting transparent obstacles such as glass, and passable tall grass in real-time, faster than existing methods that use lidars and RGB images.}}
\label{tab:inference-time}
\end{table}

Scenario 2 includes a passable string curtain which is detected as an obstacle from the 2D laser scan and occupancy map based DWA methods. Hence, these methods attempt to avoid the curtain and collide with the glass during navigation. Further, the glass detection SLAM method identifies both the curtain and the glass as obstacles.  Hence, all three methods demonstrate poor success rates in scenario 2. GDNet identified both the open door and the glass near it as glass preventing the robot from entering through the door. In contrast, our intensity map demonstrates a significantly high success rate irrespective of the lighting conditions due to the $360^{\degree}$ LiDAR-based intensity map formulation.

\begin{figure*}[t]
    \centering
    \includegraphics[width=0.95\linewidth,height=6cm]{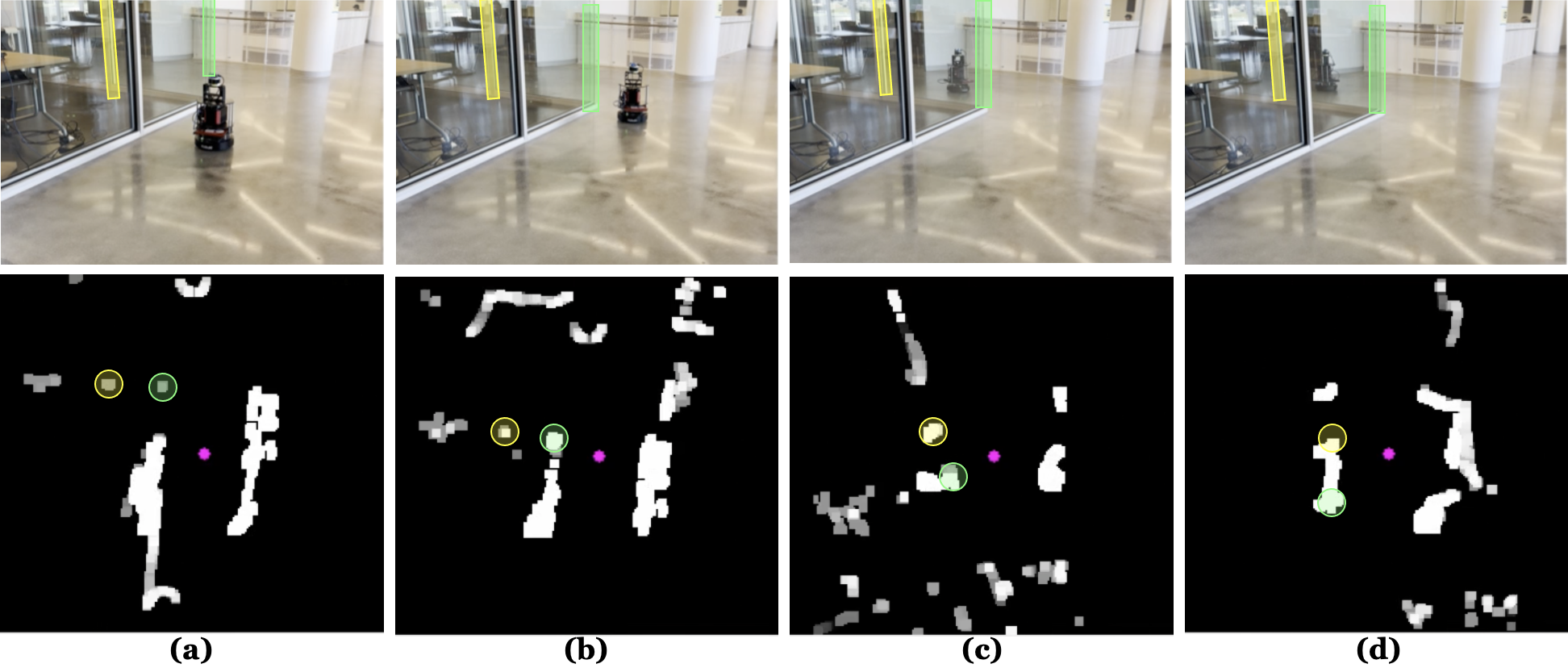}
    \caption{\small{\textbf{[Top]:} Figures depict Turtlebot's navigation using MIMs in scenario 1 with two pillars in the glass wall marked in green and yellow. \textbf{[Bottom]:} The corresponding $I^t_{plan}$ with uniform inflation. The pillars are marked in the same colors. (a) We observe that glass is misidentified as free space near the green pillar. (b) MIM's extrapolation of glass from a small neighborhood of reflected points showing up in white, as the robot moves. (c) We observe that the glass between the green and yellow pillars are misidentified as free space at a time instant. (d) The glass between the pillars is extrapolated.}}
    \label{fig:glass-extrapolation}
\end{figure*}

\begin{figure}[t]
    \centering
    \includegraphics[width=\columnwidth,height=6cm]{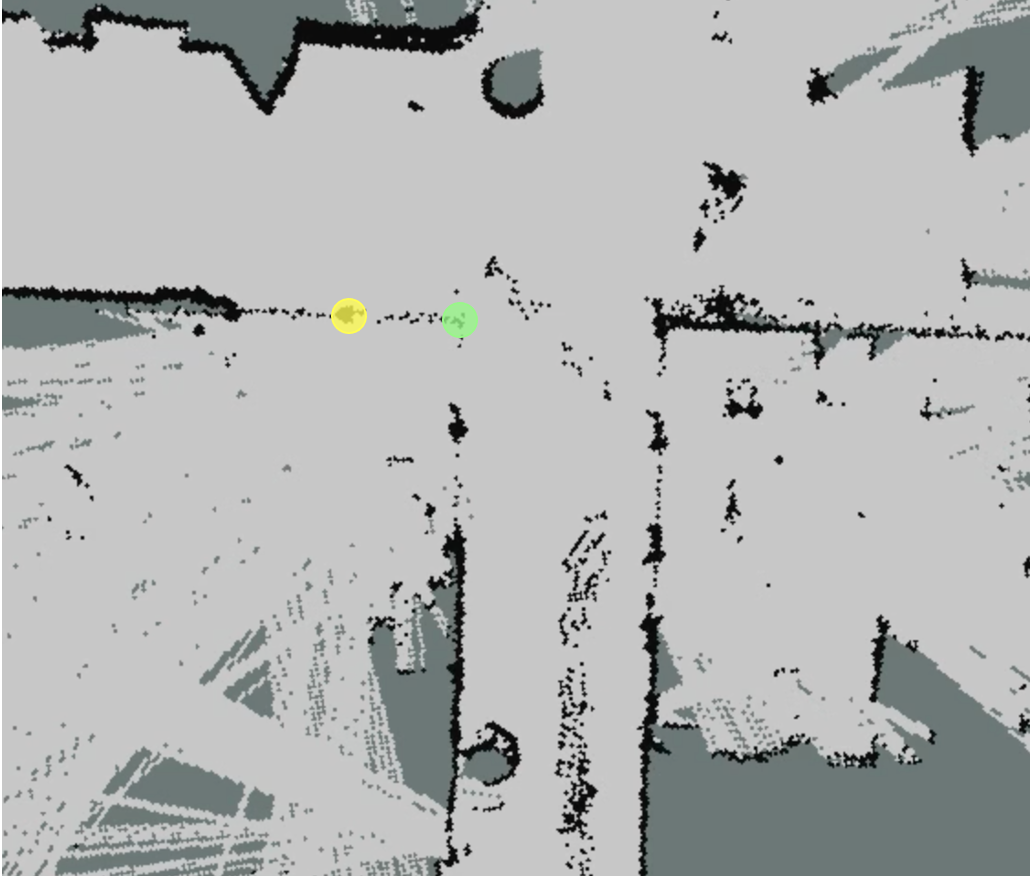}
    \caption{\small{The map created by Glass-SLAM \cite{WANG201797_glass_detection_laser_intensity} after three traversals by the robot around the glass wall. We observe that the glass is extrapolated mildly between the two pillars (see Fig. \ref{fig:glass-extrapolation}). However, since this extrapolation is not performed in real-time, it cannot be used for collision avoidance as the robot navigates an unmapped environment.}}
    \label{fig:glass-slam-results}
\end{figure}

Scenarios 3 and 4 includes traversable tall grass and non-traversable bushes and trees. However, the two DWA methods identify all such vegetation regions as obstacles due to 2D laser scan-based obstacle detection causing freezing behaviors and longer detours during navigation. Similarly, Spot's inbuilt autonomy struggles to estimate the ground and free space from its vision-based perception in both Scenario 3 and scenario 4. Hence, the robot demonstrates highly unstable motion in vegetation. In scenario 3, VERN is able to navigate through tall grass while avoiding trees and bushes using its vision-based classifier and the occupancy map formulation. However, VERN's vison-based classifier could not detect trees behind the grass region in scenario 4 since they are closely intertwined. In contrast, our multi-layer intensity map representation identifies such hidden solid objects to avoid during navigation. Hence, intensity map demonstrates a relatively higher success rate and F-score in scenario 4. In all scenarios, intensity maps (uniform and adaptive) have the highest F-score. The inaccuracies in detecting obstacles in intensity maps occur when the robot/lidar is closer than $0.5$ meters away from an obstacle, where 3D lidars typically have a blindspot.

\textbf{Benefits of Adaptive Inflation:} We use scenarios 2 and 3, which contain narrow passages to highlight the benefits of our adaptive inflation formulation. Laserscan and occupancy map-based DWA, and intensity map (Uniform) use uniform inflation around the obstacles to avoid collisions. This closes the narrow passages and represents them as obstacle regions, resulting in freezing or longer trajectories in the presence of narrow passages between the obstacles. However, our adaptive inflation preserves the narrow free spaces in the cost map while inflating the obstacles. Hence, intensity map can navigate through such spaces (e.g., through a small door in scenario 2 and between the trees in scenario 3) and reach the goals using shorter trajectories.

\textbf{Inference Time:} We compared the inference times (Table \ref{tab:inference-time}) of using intensity maps to detect glass and pliable vegetation with other methods that either detect glass (Glass-SLAM, GDNet) or vegetation (VERN) on the Intel NUC described in section \ref{sec:implementation}. We observe that apart from being versatile in detecting obstacles, intensity maps are computationally light to be used with a robot's limited onboard computing resources. GDNet and VERN use RGB images passed through deep neural networks and require extensive prior training. While Glass-SLAM does not need training, it requires $\sim 4.5$ seconds to update a map with obstacles and multiple runs to detect glass. 


%% file: 6_Conclusions.tex
\section{Conclusions, Limitations and Future Work}
We introduce a novel obstacle representation designed to enhance autonomous robot navigation in complex indoor and outdoor environments. Based on the intensity of reflected points from point clouds, intensity map effectively characterizes obstacles by their height, solidity, and opacity. Also, we present an adaptive inflation technique that further refines navigation planning by considering obstacle solidity and available free space. We demonstrate significant improvements in navigation metrics such as success rates, trajectory lengths, and F-scores, validating our proposed approach.

Our method has a few limitations. Since our multi-layer map representation is based on point cloud intensity, it cannot identify passable objects such as cloth curtains and metal fences. This is especially important because the navigability of such objects depends on the context (e.g., window curtain may not be passable but door curtains are generally passable). Hence, semantic understanding of the environment is required for such cases. Further, our method cannot detect extremely thin objects such as thin poles since 3D point cloud may not be able to capture sufficient number of samples. 
